# LiDAR Ground Filtering Algorithm for Urban Areas Using Scan Line Based Segmentation

Lei Wang, Yongjun Zhang
School of photogrammetry and remote sensing, Wuhan University, China

*Abstract*—**This paper addresses the task of separating ground points from airborne LiDAR point cloud data in urban areas. A novel ground filtering method using scan line segmentation is proposed here, which we call SLSGF. It utilizes the scan line information in LiDAR data to segment the LiDAR data. The similarity measurements are designed to make it possible to segment complex roof structures into a single segment as much as possible so the topological relationships between the roof and the ground are simpler, which will benefit the labeling process. In the labeling process, the initial ground segments are detected and a coarse to fine labeling scheme is applied. Data from ISPRS 2011 are used to test the accuracy of SLSGF; and our analytical and experimental results show that this method is computationally-efficient and noise-insensitive, thereby making a denoising process unnecessary before filtering.**

## I. Introduction

Airborne Light Detection and Ranging (LiDAR) is a kind of map survey technology emerged in the last 20 years. It has advanced dramatically during the last two decades. With the ability of direct acquiring high precision 3D point clouds efficiently, it has been widely applied in the generation of digital elevation models (DEMs) [1], [2] which is vital in water flow or mass moving monitoring, mapping and remote sensing. A typical LiDAR system is composed of a laser scanner and GPS/IMU. Mount the system on an airplane, it scans the 3D surface of the ground when flying by. The GPS/IMU is used to acquire the location and direction of each laser beam so that the 3D surface can be reconstructed precisely. Compared to traditional airborne photogrammetry methods which uses image matching techniques to reconstruct 3D ground surface, it's easier in processing(no effort is needed to generate 3D surface points) and more accurate in generated DEM. Moreover, laser pulse is able to penetrate leaves, ground under trees can also be detected. In order to generate DEM, the ground points in the LiDAR data must be distinguished from the non-ground points (e.g., trees or buildings). This process is called ground filtering [3].

LiDAR data filtering has attracted an enormous amount of research since its emergence. Many filtering algorithms have been proposed. The basic idea of all kinds of ground filtering methods is that ground and non-ground are different in local elevation and geometry characteristics such as slop, local smoothness and structure.

[Email: lei.wang@uwaterloo.ca]

The widely used TIN (Triangulated Irregular Network) based ground filtering methods ([6], [7], [8], [9], [10]) are progressive filters. A coarse terrain model is generated first. New ground points are added according to their distance and slope to the current terrain model to refine the terrain model gradually. In 2004 [11], ISPRS evaluated different filtering algorithms and the result showed that the TIN-based progressive filtering method [7] produces the best filtering results and adaptation ability. Instead of TIN, Mongus 2012 [12] used thin plate spline as the ground model in iterative filtering. The author achieved non-parametric filtering, but the local lowest points were selected as control points, which means the filtering results must rely on the denoising algorithm. Belkhouche 2011 [13] tried to classify the edge triangles from TIN using slope, and only the non-edge triangles remained. These methods can adapt to different topologies due to their attempt to compare with the local ground model. Further, these algorithms are not noise-resistant. Errors accumulate easily through the progressive process. Slop based filtering is a classical ground filtering method proposed by Vosselman [3]. The local maximum slope is evaluated from point clouds and been used to discriminate ground points and non-ground points. The radius and the slope threshold need to be designated beforehand. Normally, the maximum slope in a test area is used as the slope threshold, and the radius is slightly larger than the building size. Later, Sithole proposed an improved method that calculates adaptive slop threshold from local point clouds [5]. Morphology based method was first introduced in 1996 by Kilian [14]. It performs an open operation in a local window to obtain ground points. In 2003, Zhang et al. introduced progressive window sizes to the morphological filter [15]. In 2005 Shan et al. proposed a bidirectional labeling method [16]. It generates a profile first and tries to extract the ground point by the slope and elevation difference together with the classes of its neighbors. This method can effectively detect small objects such as cars and shrubs. Meng et al. (2009) proposed a multi-directional ground filtering (MGF) method [17] that makes use of information along the scan line and across the scan line. The filtering results are not affected by the building size; but the slope threshold and elevation difference threshold need to be designated according to the topology. Also, the ground seed point needs must be determined, and the method is noise-sensitive. Haala and Brenner (1999) [18] incorporates optical image into LiDAR point cloud, then uses ISODATA to classify the ground. This method requires precise registration of the LiDAR data and images. For high resolution images, classification is a very tough problem. In 2002, Felin proposed a segmentation-based method that combines point location, elevation difference with neighbors, and relationship with the tangent plane to complete segmentation and classification [19]. In 2003, Jacobsen used Ecognition to conduct the segmentation, and the elevation difference was used for classification [20]. Tovari used region growing in 2005 to do segmentation, and an iterative weighted local plane fitting algorithm was used to classify the ground points [21]. Rabbania proposed a segmentation method based on smooth regularization in 2006 [22]. These LiDAR data segmentation methods can be divided to two types: (a) plane fitting-based and (b) local feature similarity-based. The plane fitting-based method uses the plane to describe the shape of the local ground, which is not always a correct assumption. This method also cannot adapt to different

---

[Email: lei.wang@uwaterloo.ca]

topologies while the currently used local feature similarity-based segmentation methods use the local elevation difference, angle difference, local flatness, etc. to build segments. These two different methods both select points within a bounding box as the segmentation target, but this segmentation-based method makes small targets hard to detect, so iteration and post-process are needed. Another problem is that even though each plane is separated, the topological relation is complex, especially in complex roofs, which makes the filtering process difficult.

As can be seem, different ground filtering methods explore ground/non-ground differences in a) different strategies, b) data processing elements/neighbor definitions, c) features, and d) parameter estimation methods. For strategy, there are progressive and non-progressive filtering methods. Progressive methods try to refine the filtering result by repeating the same filtering process multiple times. Non-progressive methods just run one circle. Progressive methods are able to adapt to different local topology types but the error is also propagated with the progressive process. For data processing elements/neighbor definitions, there are point based, scan line based, segment based methods. Point based methods treat each single point as a processing element, to explore the local structure, k-nearest neighbors with distance constrain is widely used as neighborhood definition. Data indexes are needed for efficiently searching neighbors. A scan line is a data profile that lies along the scan direction of laser scanner. Its neighbors are confined in the same profile, therefore, calculation is simplified than point based methods, but scan lines can only capture topology characteristics along the scan line. Segment based methods treat each segment as a basic unit. Spectral and geometric characteristics are used to obtain segments. Points belong to one segment are considered the same class. Scan line and segment based methods are noise-resistant compared to point based methods. Denoising is necessary before applying point based method which is a highly researched topic. While current segmentation methods are sensitive to parameters. Features has been used include slop, height, local flatness, local fluency, other dataset (images, GIS dataset). Common parameters such as threshold of height difference, threshold of slop difference and window size can be predefined or estimated from neighbors, or local ground model such as local fitted spline or TIN. Parameter estimation is essential to a filtering algorithm. Estimating parameters from neighbors or local ground models is called adaptive method. It's more flexible than uniform threshold.

These algorithms work well in flat areas with small vegetation coverage. Due to the complexity of the urban land topology (mainly cliffs, small bushes, large buildings), these ground filtering methods can't generate satisfying results. Several aspects still need to be improved: noise resistance, robustness, threshold sensitivity. Computation efficiency is another important issue as well. Data volume is growing dramatically and real-time analysis is required for many applications, such as disaster rescue. These reasons inspired us to develop a new ground filtering method that is robust to noise and topology conditions, less sensitive to parameters, and fast.

The contribution of this paper is that it proposes a ground filtering optimized LiDAR point cloud segmentation method using nature scan lines. Compared to current scan line segmentation methods which re-sample the original data into a grid-aligned form

[Email: lei.wang@uwaterloo.ca]

to generate regular scan lines, our method uses the original scan lines obtained in data acquisition. The segmentation algorithm is designed to segment the same type of point (ground/non-ground) to the same region as much as possible in order to simplify the topologies between regions. For example, different sides of the same roof are connected to the same region. Therefore, it can facilitate the filtering process. It is computationally very efficient and easy to use. We also designed an iterative filtering algorithm based on the scan lines and segmentation results. The basic idea is similar to the TIN-based filtering methods, except that the whole filtering operation is based on scan line segments and regions. It's more efficient than point based progressive filtering methods.

The structure of the remainder of this paper is as follows. In Part II, the filtering method will be introduced in detail. The filtering experiment and results are presented in Part III. A comparison between the proposed method and the Terrascan's TIN-based filtering method is also made in Part III. Part IV is discussion and conclusion.

## II. Ground Filtering using Scan Line Segmentation

Our proposed method is essentially a segmentation-based method. The overall algorithm is composed of three steps.

1) Extract scan lines and extract homogeneous continuous point sequences which we call segments from each scan line.
2) Assess the similarity of the neighboring segments and connect adjacent similar segments to form regions (segments).
3) Conduct progressive filtering based on scan lines and regions; select initial ground segments and build initial DTM; iteratively update the current DTM with the current labeled "ground" regions; label the qualified regions as ground based on the current DTM until stable.

### *A. Extract continuous segments in scan lines*

The detailed work flow is as below:

1) Extract and rearrange the scan lines using the scan direction and the indicator of end-of-scan-line in each point record. There may have small back scanning while data acquisition due to the rolling of airplane. These back scanning points can be easily detected and deleted by check the coordinate difference along scan direction. Then, rearrange the remaining points in each scan line along the scanning direction.
2) Calculate slope difference *Dslope*$_i$ for each point *i* along scan line direction use equation

$$\begin{aligned} Dslope_i = {} & Z_{i+1} - Z_i \,/\, \sqrt{\left(X_{i+1} - X_i\right)^2 + \left(Y_{i+1} - Y_i\right)^2} \\ & - Z_i - Z_{i-1} \,/\, \sqrt{\left(X_i - X_{i-1}\right)^2 + \left(Y_i - Y_{i-1}\right)^2} \end{aligned} \quad (1)$$

   Where $\left(X_{i+1}, Y_{i+1}, Z_{i+1}\right)$, $\left(X_i, Y_i, Z_i\right)$ and $\left(X_{i-1}, Y_{i-1}, Z_{i-1}\right)$ represent the three dimensional coordinates of three sequential points in a scan line.
3) For each point *i*, if *Dslope*$_i$ > *Tdslope*, delete this point. A sequence of points is therefore obtained. Continuous point sequences

[Email: lei.wang@uwaterloo.ca]

with no “holes” are extracted, only the two end points of a sequence are kept and connected, and the scan line is then split to separate segments.

4) Delete short segments so that trees and small objects will be erased.

In point cloud dataset gathered by LiDAR systems, there are a scan direction indicator and scan line end indicator for each point. Points that belong to the same scan line can be extracted using these two indicators. The ground or objects on the ground might change along arbitrary direction, so it is not necessary to generate an artificial scan line by slicing the point cloud to profiles as has been done in [16, 17]. The extracted scan line is a sequence of points arranged by point coordinate along the perpendicular direction of the scan direction.

For both grounds and roofs, the slope differences are relatively small and stable; but at the edges of ground or roofs (walls), slope differences change dramatically. Trees and shrubs have the same characteristics as edges due to their rough surfaces and the penetrated points under the vegetation cover. Noises behave in a way similar to very small objects and can be treated likewise. By doing this, trees and small objects can be detected and deleted. Consequently, only roofs and ground segments are remained. It will facilitate the following filtering.

*B. Segmentation using line segments*

In this step, the similarity between neighboring segments (along both the scan line direction and its perpendicular direction) is evaluated. A similarity graph is built to record the similarities between each segment and its neighboring segments. Each segment is a node of the graph. The similarities of neighboring segments are evaluated by combining their distance $D_X$ along the scan line direction, the elevation difference $\Delta h$ and the slope difference in degree $\theta Z$. Only when all these three similarity measurement are within threshold, the two neighboring segments are thought to be similar and are connected with an edge in the similarity graph. To any two neighboring segments $A$ and $B$, assume corresponding endpoints are $AS$, $AE$, $BS$, $BE$, then

$$\begin{aligned} D_X &= \min \sqrt{(b_X - a_X)^2 + (b_Y - a_Y)^2} \qquad \forall a \in A, \forall b \in B \\ \Delta h &= \min | b_Z - a_Z | \qquad \forall a \in A, \forall b \in B \\ \theta Z &= \arctan \frac{BE_Z - BS_Z}{BE_Y - BS_Y} - \arctan \frac{AE_Z - AS_Z}{AE_Y - AS_Y} \end{aligned} \tag{2}$$

The corresponding thresholds for these three measurements are $T_{DX}$, $T_{\Delta h}$ and $T_{\theta Z}$. $T_{DX}$ should be small enough to ensure that the neighboring segments are highly spatially-correlated. $\Delta h$ is the minimal elevation difference of the points between two segments. For cliffs or roof edges, $T_{\Delta h}$ will be large; While within roofs or ground, $T_{\Delta h}$ will be small and usually will be zero. So it is threshold insensitive. $\theta Z$ is the slope difference of the neighboring segments in different scan lines. In order to segment a complex roof structure into one segment, $T_{\theta Z}$ should be large enough. Moreover, $\theta Z$ can help detect the segments on walls for the slopes are close to $90^{\circ}$ in such situations. So the value of $T_{\theta Z}$ can be very flexible. A region is consist of a set of segments that each node is connected directly or indirectly with the rest segments in that set. Regions are detected by an ordinary traversal algorithm. In our implementation, a depth-first traversal is used.

### *C. Progressive filtering based on scan lines and regions*

Like all other progressive filtering methods, an initial coarse DTM first is obtained, then iterative refining is applied to optimize the initial DTM. To build an initial DTM, certain local lowest line segments are labeled as ground using a rigorous threshold. Similarly, some of the highest line segments are labeled as non-ground. After a consistency adjustment and segment adjustment, initial ground line segments are obtained. The initial DTM is then built using these points; and the remaining unlabeled line segments mainly are located in ladder topologies. The searching threshold is then relaxed and the labeling process is repeated, after which the DTM is refined. The process continues until the DTM becomes stable (doesn't change). This method is able to detect independent houses, houses with complex roof structures, and buildings with skirt buildings. The results are not affected by the size of buildings. The detailed flow chart is shown in Fig. 1.

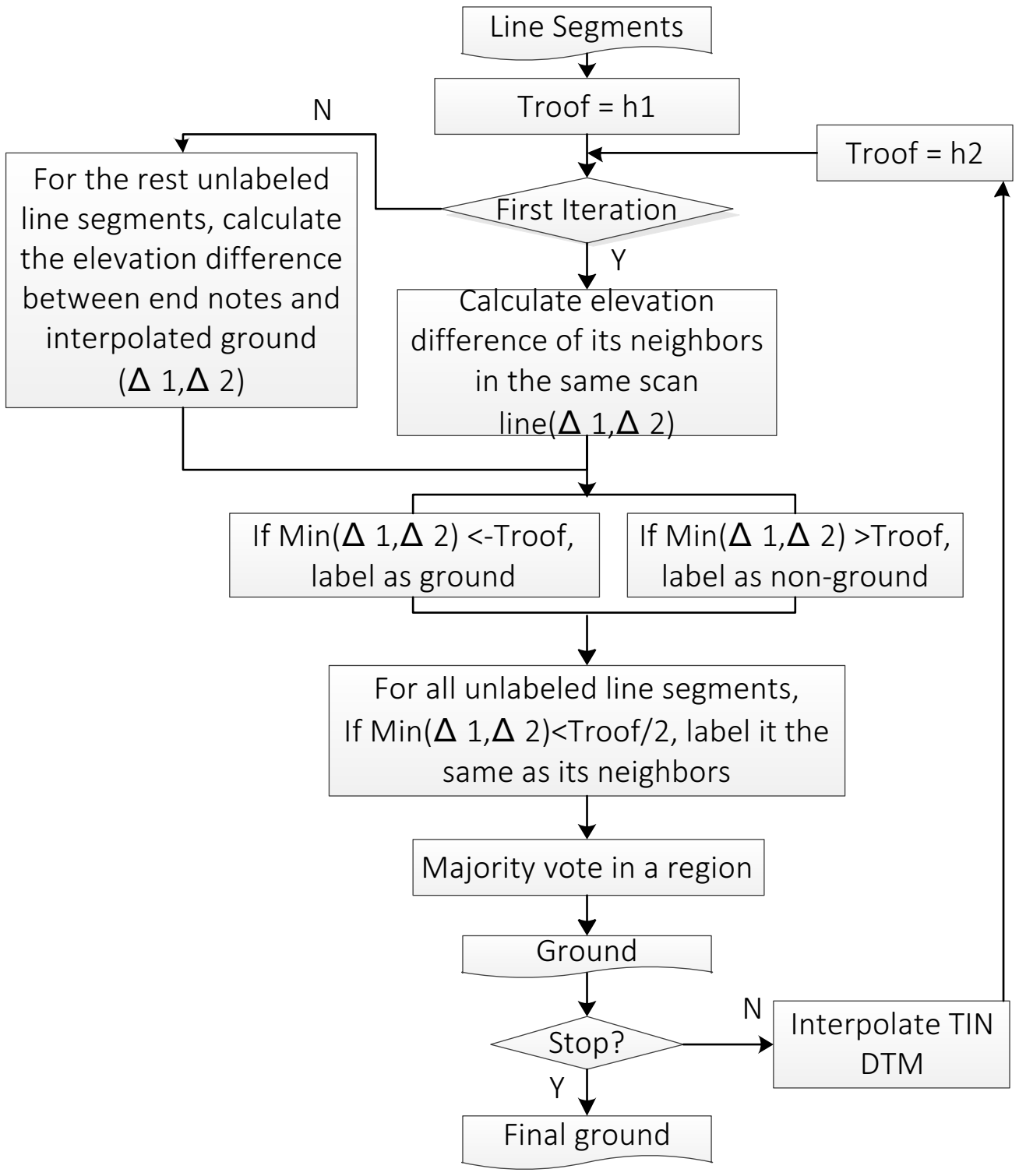

Fig. 1. Work flow of progressive filtering based on scan lines and regions.

## III. RESULTS AND DISCUSSION

The test area selected for our experiment is located at Vaihingen City in Germany which is a fairly well developed city. The test dataset is provided by an ISPRS 3D reconstruction project ([23], [24]). The dataset was acquired by the Leica ALS50 laser scanner, with FOV *45*°, flight height about *500* meters. Average footprint density is approximately *4points/m*$^2$. The dataset extends *1,450m*

[Email: lei.wang@uwaterloo.ca]

in the east-west direction and *350m* in the north-south direction (Fig. 2). The test region covers four common land forms in cities:

1) Hills (Fig. 2(b)): located at the west boundary of Vaihingen, covered by vegetation and buildings.
2) Residential area (Fig. 2(c)): distributed in the whole test area, some front yards are several meters higher than the road with many small surrounding objects.
3) Commercial district (Fig. 2(d)): located in the middle of Vaihingen, complex buildings and roofs are its main characteristic, and there are trees along the streets (Fig. 2(d)).
4) Cliffs (Fig. 2(e)): most cliffs are located around residences, along the river and on the mountain.

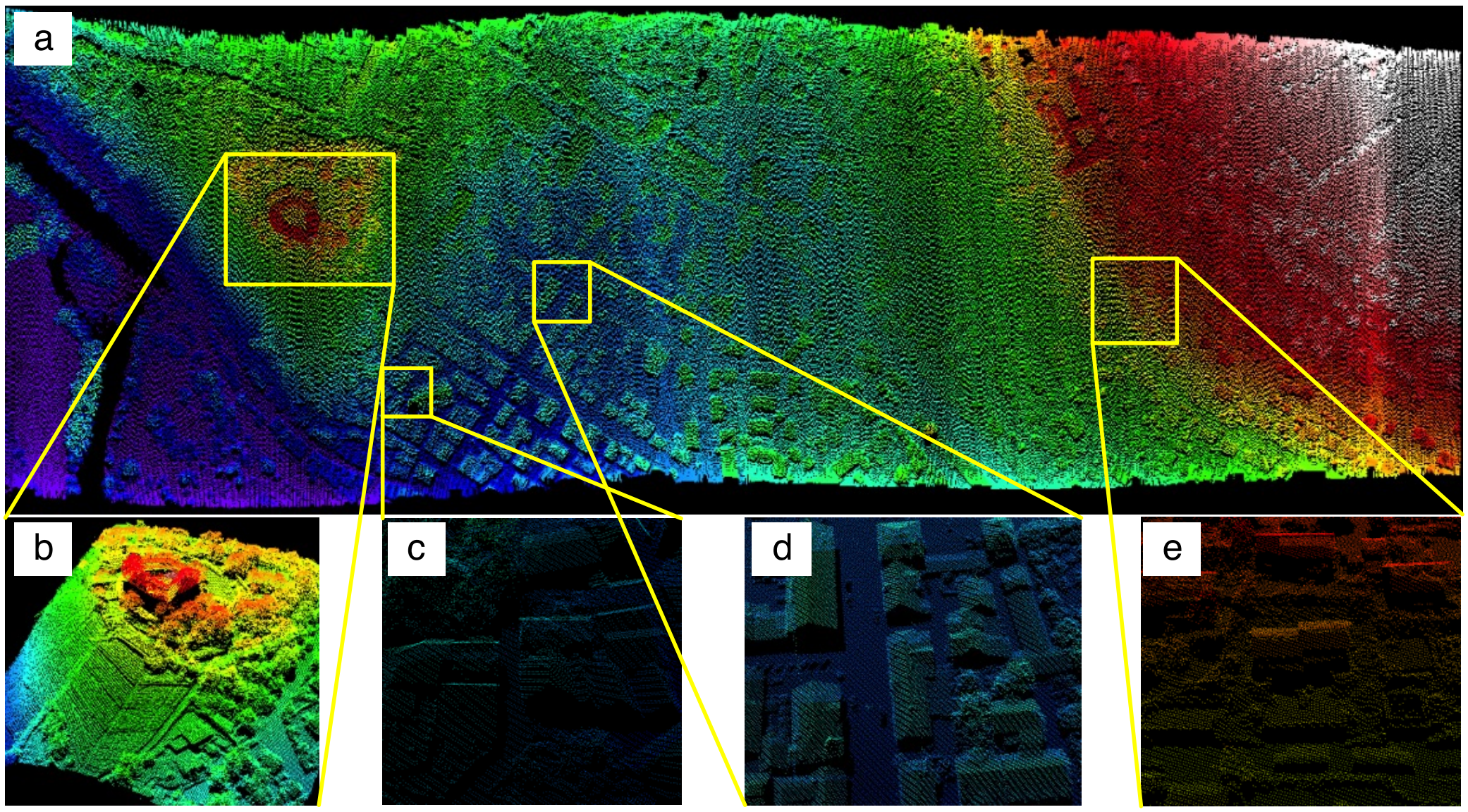

Fig. 2. Vaihingen test area. (a) is the overview of the entire test area, (b, c, d, e) are different topology types.

Set the slope difference threshold *Tdslope=0.5* for the scan line segment, the distance threshold of line segments $T_{DX}=1m$, the height difference threshold $T_{\Delta h}=0.5$, and the angle difference threshold $T_{\theta Z}=30^{\circ}$ for segmentation. The segmentation results are shown in Fig. 3. Blocks of the same color that are separated from their surroundings are regions. Points that do not belong to any regions (deleted points in the line section process) are displayed in different colors according to their slop differences along scan line direction. As can be seen in the results, the roofs are well detected and the most complex roofs are segmented as one region. Also, the walls are fully detected and erased. Because ground are generally continuous, the ground segments are large in this area. Most of the small ground objects, such as shrubs or cars, are removed, and at the end of the segmentation process, only the ground and roofs are remained in the original data.

[Email: lei.wang@uwaterloo.ca]

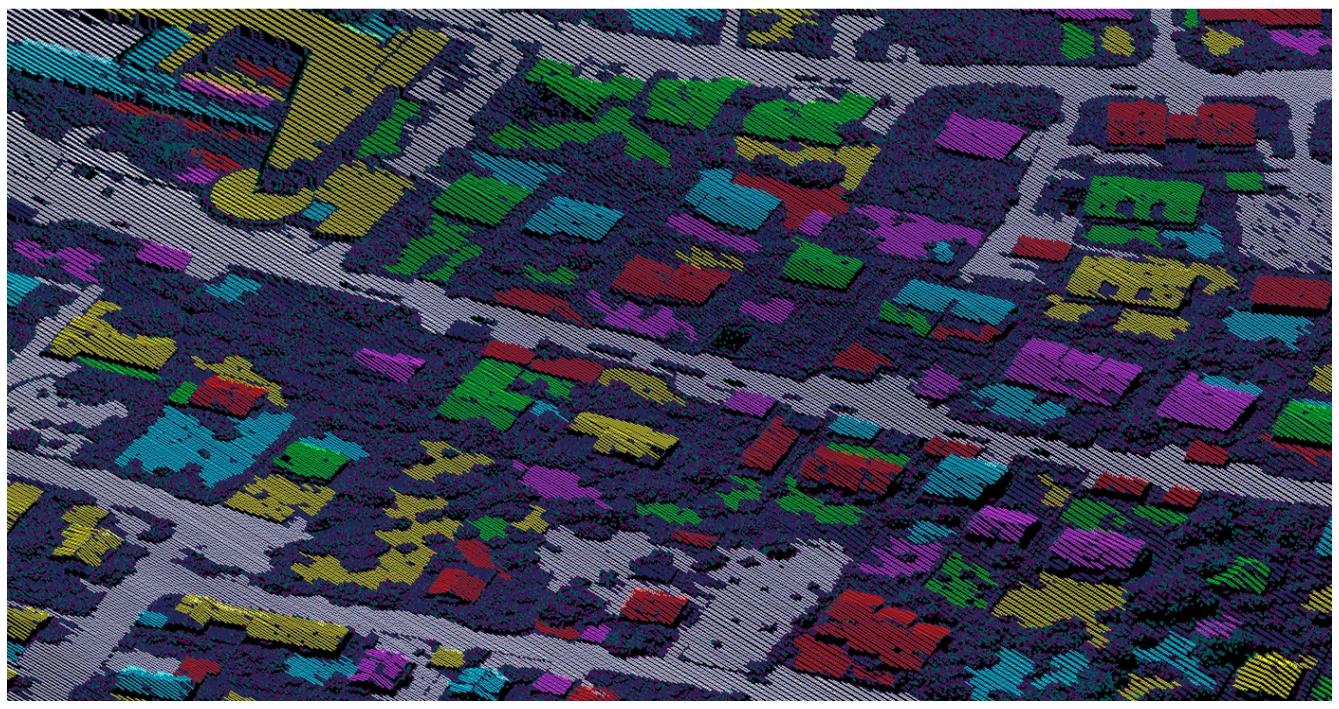

Fig. 3. Part of the segmentation result.

In the iterative labeling step, *h1* is set to *0.5m* to obtain the initial ground line segments, and *h2* is set to *2m* in the following iteration. The final filtering result is shown in Fig. 4. When compared to ground truth that generated by manually interpretation, the type I and type II errors are *2.2%* and *2.1%* separately in segmented areas. Because the ground points under vegetation are not detected in this proposed algorithm, the true filtering error percentage should be larger than shown. The ground under vegetations is estimated by interpolation. A great deal of the mislabeling happens with small ground objects, such as cars and trucks, and this mislabeling is one of the main drawbacks of this method. The detailed filtering results for the different topology conditions are shown in Fig. 5. In the hill areas, due to the dense vegetation coverage, the open ground area is relatively small; therefore, the DTM interpolation, which is based on a few ground patches, does not work well and an arrow-pointed cliff area also is left out. In the downtown area, the ground tends to be flat and there is less vegetation cover so small objects are deleted. In the residential area, some of the cliffs are also missing due in part to the cliffs being relatively tall (over two meters) as well as being surrounded by vegetation so the interpolated ground model under the vegetation contains more errors.

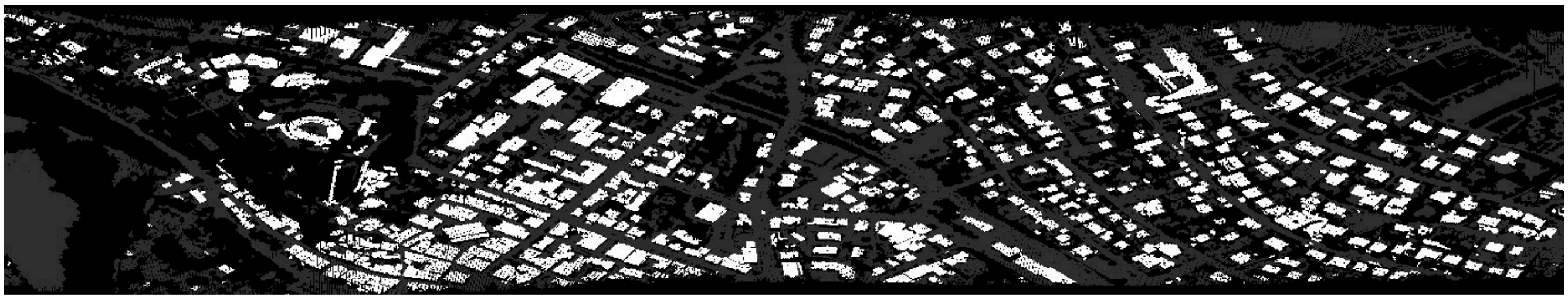

Fig. 4. Filtering result using $T_{roof} = 2m$.

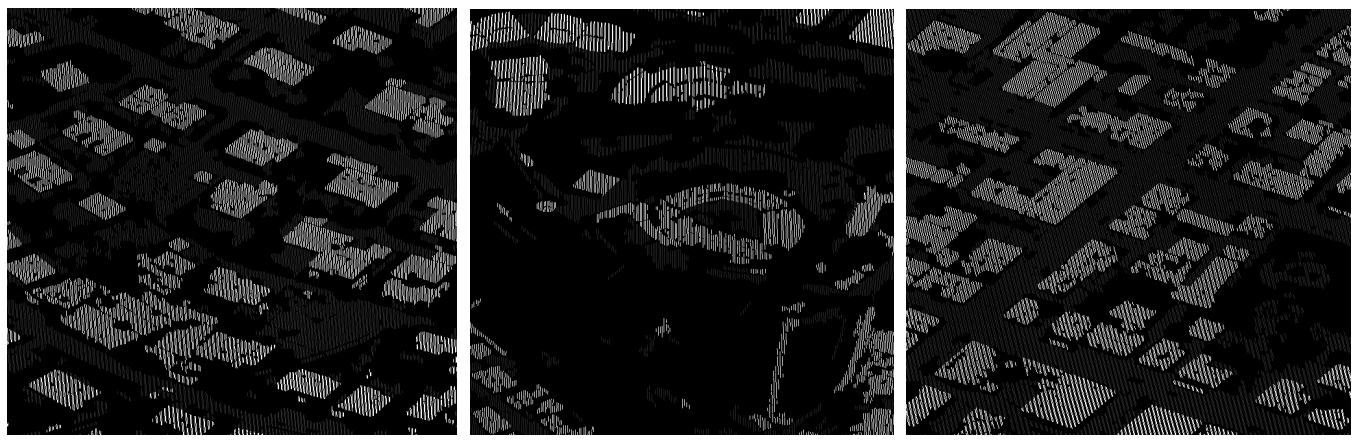

(a) Hill (b) Residence (c) Cliff

Fig. 5. Details of filtering result in difference topology conditions.

To test the accuracy of our method, the ground filtering module of Terrascan, a well-known commercial LiDAR data processing software, was used to conduct a comparison. Terrascan uses a progressive TIN interpolation filtering algorithm proposed by [5]. In this experiment, the maximum building size is about *60m* which is manually measured in Google Earth. The maximum slope is $88^o$

[Email: lei.wang@uwaterloo.ca]

which is calculated from the dataset, the angle threshold in iteration is *6°* which is the default value of Terrascan, the distance threshold is 1.4m. Terrascan's filtering results overlaid on our results are shown in Fig. 6. *12.7%* of the total points are labeled as ground by our method and as non-ground by Terrascan. These points are composed of four kinds: large areas of ground points missed by Terrascan (Fig. 7(a)), ground points on the edges of cliffs (Fig. 7(b)), small and low roofs mislabeled by SLSGF (Fig. 7(c)), and sparse ground points missed by Terrascan (Fig. 7(c)). It is believed that the cliff was missed by Terrascan because the distance between the cliff edge points and the triangle in its area is normally larger than the points that are located away from the cliff. Also the missed sparse ground points by Terrascan are not errors, but rather were caused by Terrascan's TIN vertex density control. Terrascan's ground and SLSGF's non-ground were mainly distributed in the cliffs (Fig. 7(d)), and both results contained errors. For example, for SLSFG, vegetation and small object points that did not belong to any segment were labeled as "Unclassified"; and since it increases the interpolation errors in hills, therefore high cliffs were missed by SLSGF (Fig. 7(g)). As for Terrascan, it missed some ground around the woods and inside the building (Fig. 7(e,f)).

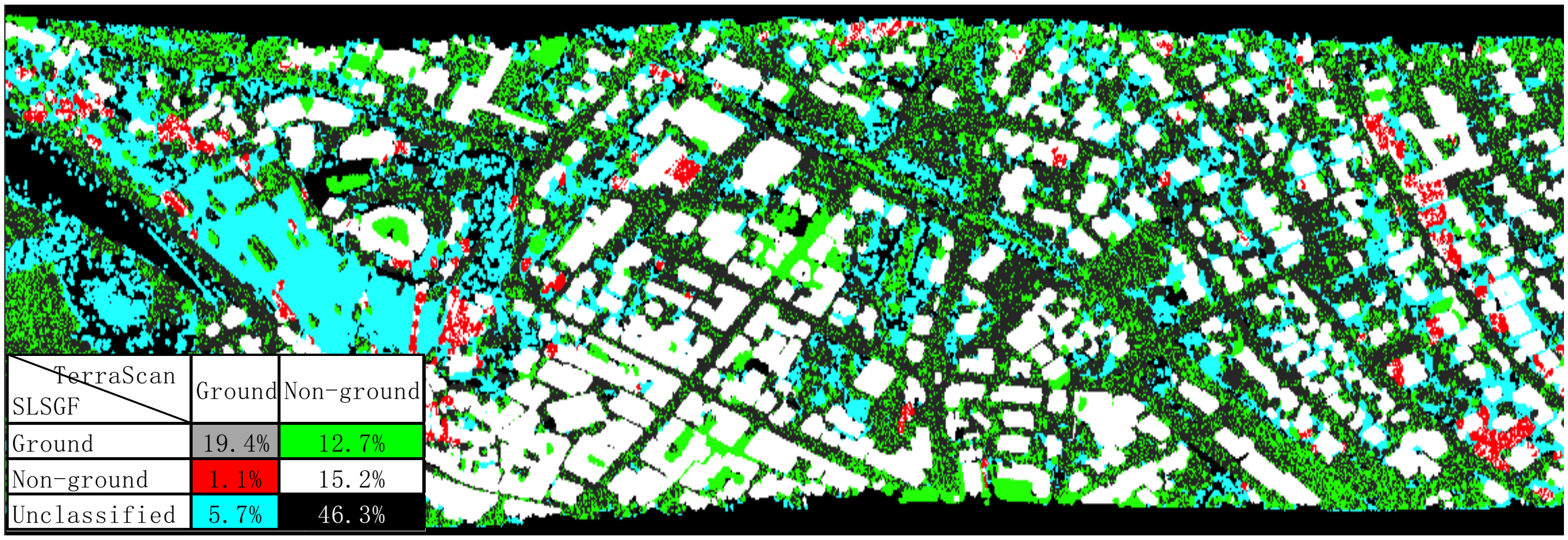

Fig. 6: Overlaid image of filtering results by Terrascan and our algorithm. Different colors correspond to different filtering result pairs in the table, for example, the red points are those labeled as ground by Terrascan while labeled as non-ground by our algorithm. The values in the table are the percentages of the corresponding situations in all points.

[Email: lei.wang@uwaterloo.ca]

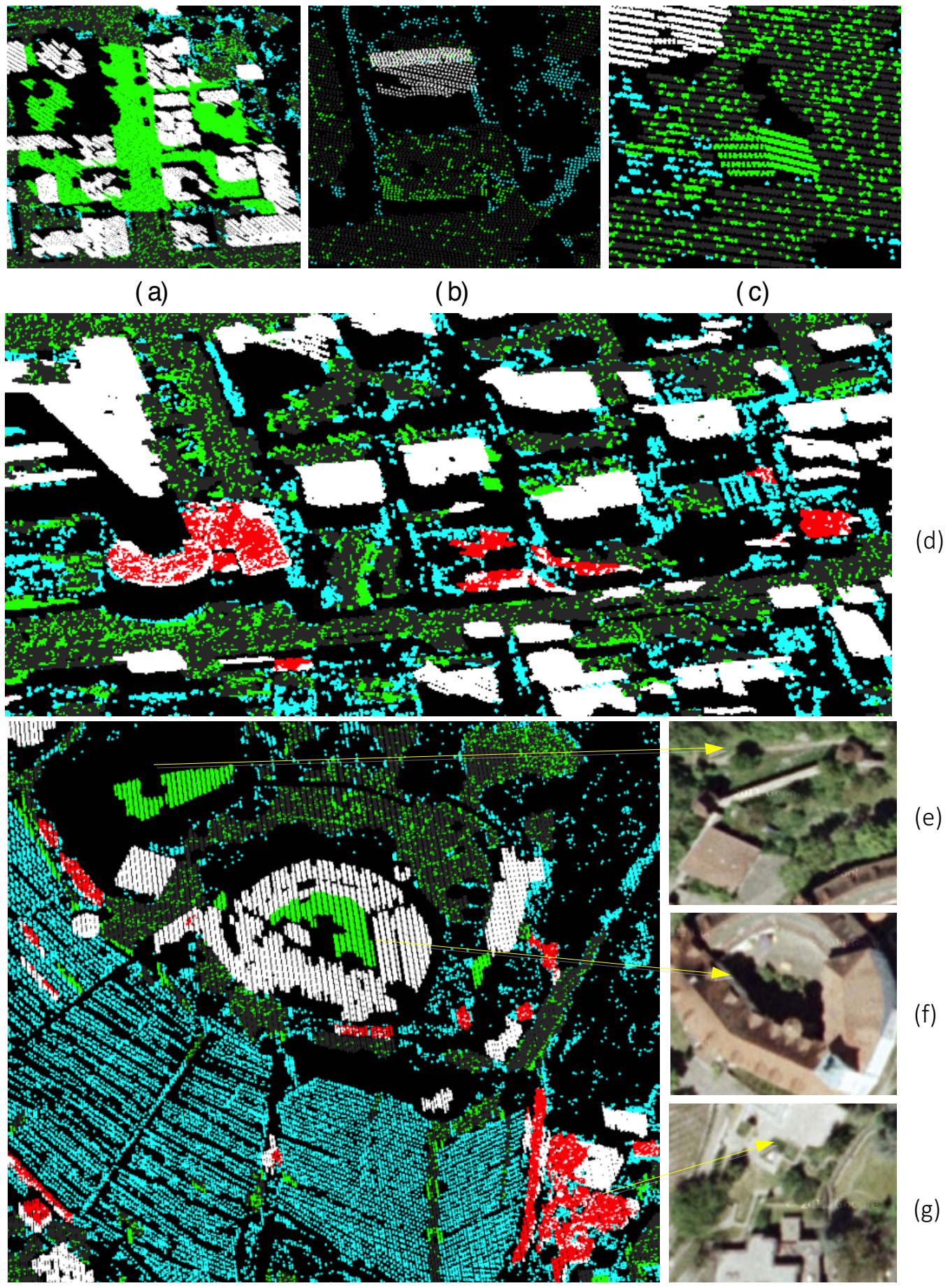

Fig. 7: Comparison of SLSGF and Terrascan.

The gross error resistance of SLSFG is demonstrated by comparing it to the Terrascan TIN-based filtering algorithm (Fig. 8). No obvious gross error remained for SLSFG, while many gross points remained for Terrascan.

In order to test parameter sensitivity, segmentation was done at a different *Tdslope*. The changes in the number of line segments and final segments with respect to *Tdslope* are shown in Fig. 9. Generally, when *Tdslope* is relaxed (increased), the number of line segments increases as well, but becomes stable when *Tdslope* is larger than *0.2*. When *Tdslope* is *0.1*, which is too low, the segmented lines are not good enough to compose meaningful segments. When *Tdslope* is *0.2*, the segments are small and the roofs are separated into different segments; and as *Tdslope* continues to increase, the roofs gradually combine into single segments and the segment results therefore becomes stable. So when *Tdslope* changes by a certain amount (around *0.5)*, the segmentation results experience only small fluctuations.

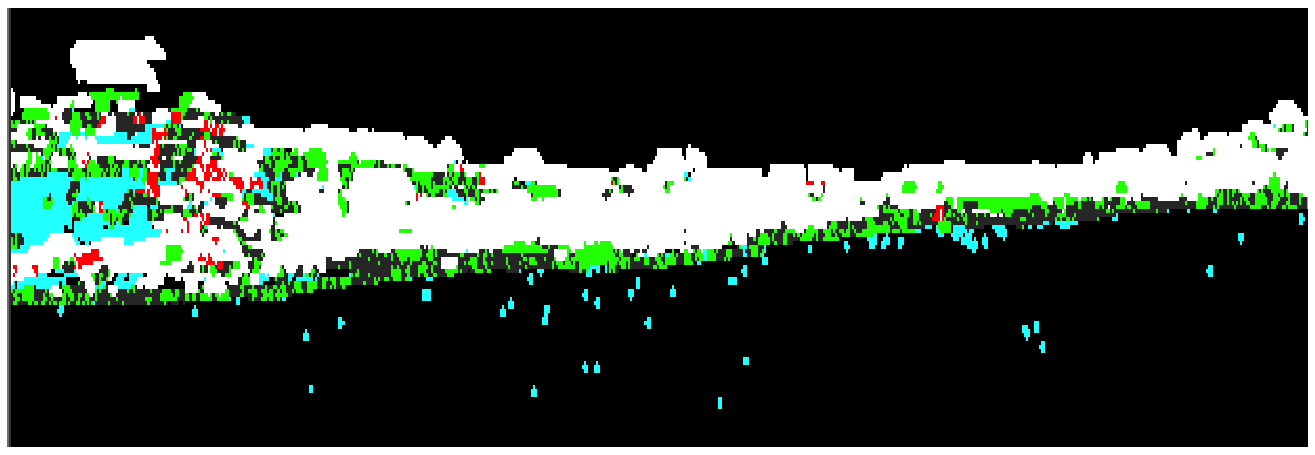

Fig. 8: Remained gross errors by TIN-based filtering algorithm with Terrascan.

[Email: lei.wang@uwaterloo.ca]

Assuming that the horizon distances between point pairs *(i+1,i),(i,i−1)* are a and b separately, the second-order derivative can be written as:

$$Dslope_i = \frac{Z_{i+1} - Z_i}{a} - \frac{Z_i - Z_{i-1}}{b} = \frac{1}{a}Z_{i+1} - (\frac{1}{a} + \frac{1}{b})Z_i + \frac{1}{b}Z_{i-1} \quad (3)$$

For three sequential points in a scan line, $a \approx b \approx dpoints$, *dpoints* is the average point distance at its location. When no gross error exists, the calculated *Dslope* has standard deviation *σDslope* (4).

$$\sigma_{Dslop} = \frac{2\sigma_Z}{\sqrt{d_{points}}} \quad (4)$$

In (4), *σZ* represents the point elevation measurement precision. It can be estimated by selecting the flat roof points and calculating their standard deviations. In our experiment, the estimated value of σZ is about *0.05m*. The nadir point spacing is *0.4m*. So according to equation 4, σ*Dslope = 0.16*. Therefore, *95%* of the points of the flat areas should have second-order derivatives of less than *2σDslope*, which is 0.32. In our experiment, *Tdslope* therefore should be over *0.32* to ensure that most of the plane features are extracted.

Similarly, how the segmentation results change with respect to $T_{\Delta h}$ and $T_{\theta Z}$ are plotted in Fig. 10 and Fig. 11. When these parameters change around an optimal value, the segmentation results remain stable.

[Email: lei.wang@uwaterloo.ca]

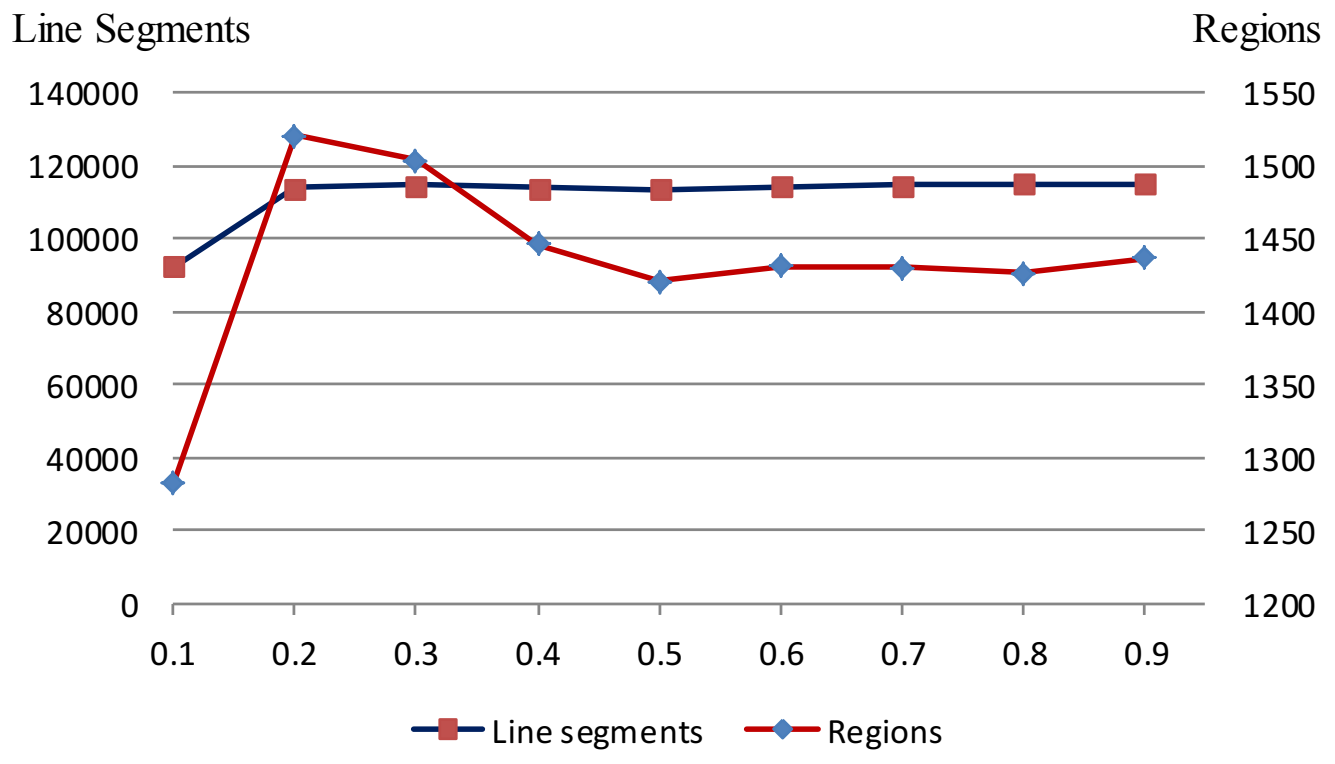

Fig. 9. Tdslope's effect on segmentation.

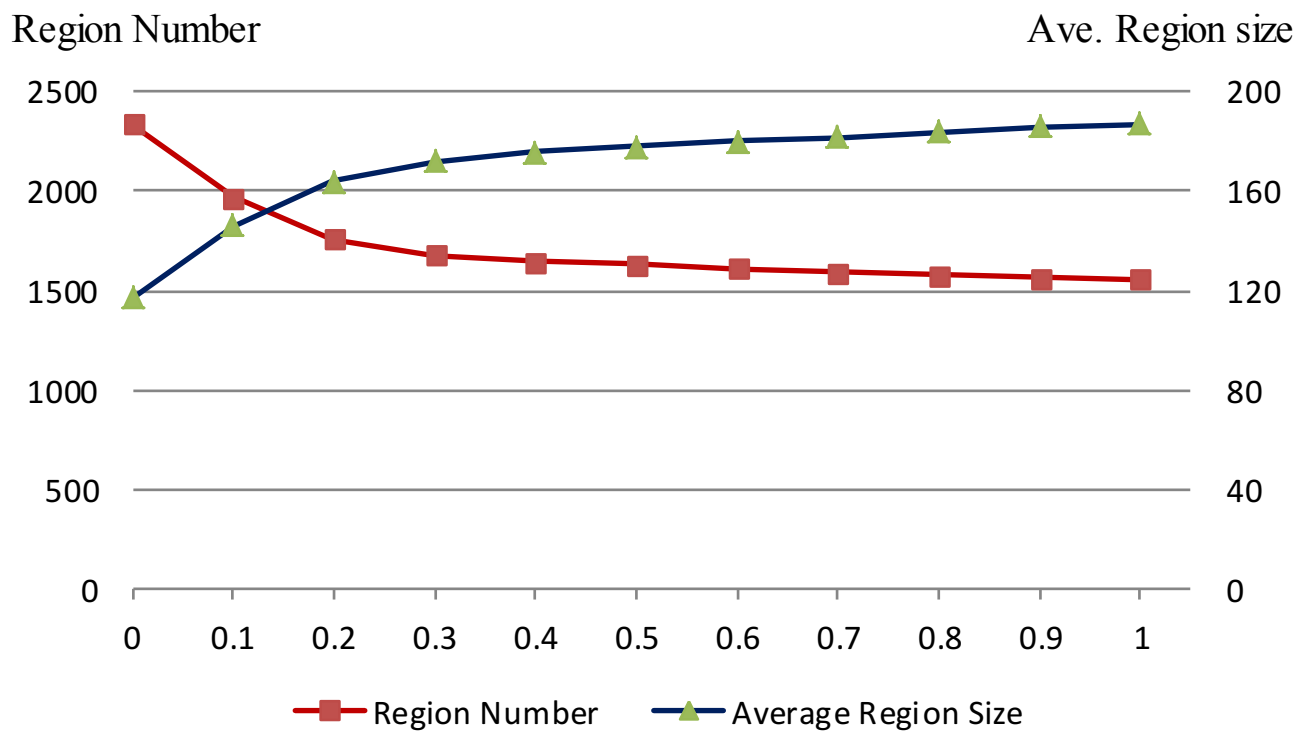

Fig. 10: Segmentation result changes with respect to $T_{\Delta H.}$

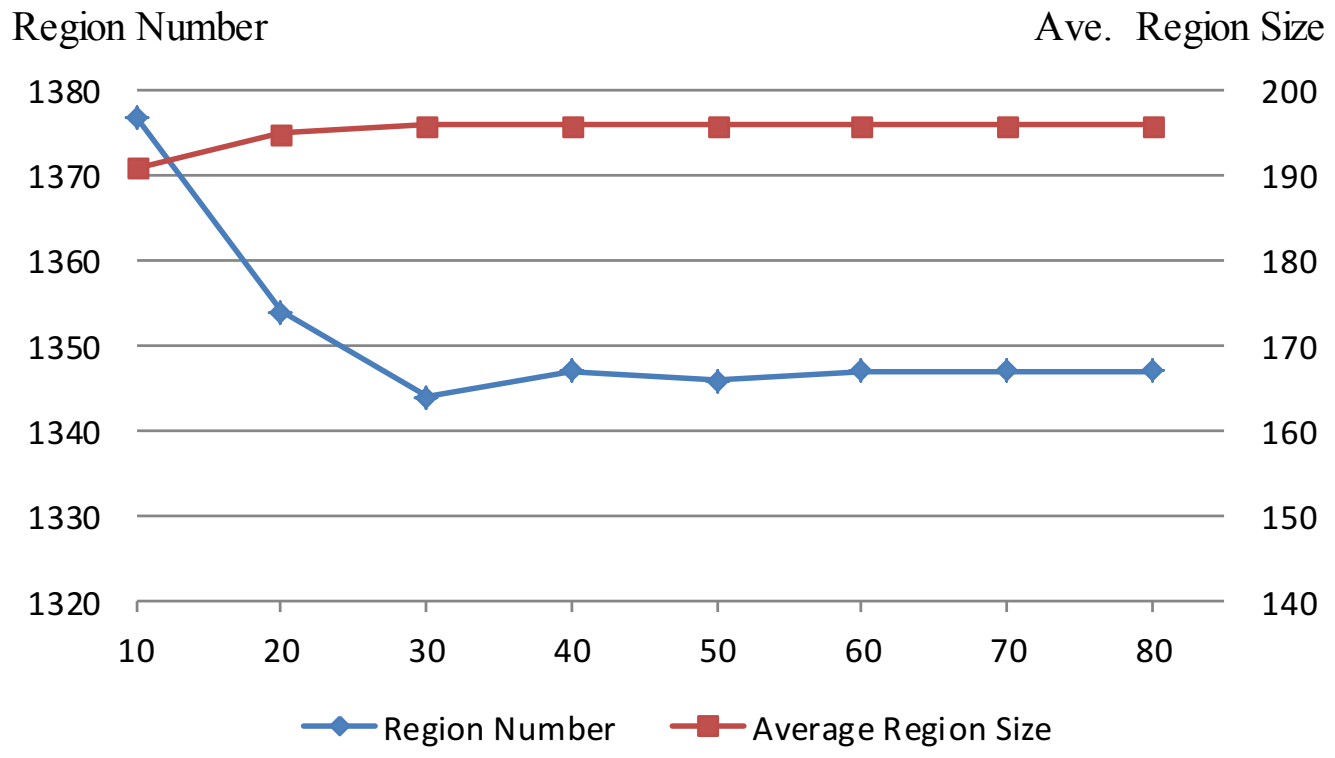

Fig. 11: Segmentation result changes with respect to $T_{Z.}$

## IV. Discussion and Summary

This paper proposed a simple but efficient LiDAR data segmentation algorithm optimized for ground filtering purposes. There is no need to build artificial scan lines, build a data index, search the k-nearest neighbors, and fit the local plane. It can produce satisfying results in urban areas, and its parameters are insensitive and robust to noise. Our work shows that scan line information can be used to facilitate the filtering process dramatically. However, this method cannot extract ground points that lie under vegetation; and when a vegetation area is large, the interpolation precision of DTM will decrease. Without a sufficiently precise

[Email: lei.wang@uwaterloo.ca]

DTM, the filtering process therefore can be undermined, and future research should address how to use low SNR under-vegetation points to estimate DTMs. Some of the parameters are derived from data statistics; and because of SLSGF's insensitivity to parameters, it is possible to build a parameter-free ground filtering method based on SLSGF. This segmentation method also can be applied to roof reconstruction and adjusted to fit other applications.

## REFERENCES

[1] G. Mandlburger, C. Briese, and V. N. Pfeifer, "Progress in LiDAR sensor technology chance and challenge for DTM generation and data administration," *In In Proceedings of 51st Photogrammetric Week 2007*, Stuttgart, Germany, pp. 159-169, 2007.

[2] N. Pfeifer, P. Stadler, and C. Briese, "Derivation of digital terrain models in the scop++ environment," *In Proceedings of OEEPE Workshop on Airborne Laser Scanning and Interferometric SAR for Digital Elevation Models*, Stockholm, Sweden, 2001.

[3] G. Vosselman, "Slope based filtering of laser altimetry data[A]. *Int. Arch. Photogramm. Remote Sens. Spatial Inf. Sci.*," Amsterdam, the Netherlands, volume 33 of part B3/2, pp. 935-942, 2000.

[4] X. Meng, N. Currit, and K. Zhao, "Ground filtering algorithms for airborne LiDAR data: A review of critical issues," *Remote Sensing*, no. 2, pp. 833-860, 2010.

[5] G. Sithole, "Filtering of laser altimetry data using a slope adaptive filter," *Int. Arch. Photogramm. Remote Sens. Spatial Inf. Sci.*, 34(3/W4), Annapolis, ML, pp. 203-210, 2001.

[6] H. S. Lee, and N. H. Younan, "DTM extraction of LiDAR returns via adaptive processing," *IEEE Trans. Geosci. Remote Sens.*, vol. 41, pp. 2063-2069, 2003.

[7] P. Axelsson, "Processing of laser scanner data-algorithms and applications," *ISPRS J. Photogramm. Remote Sens.*, vol. 54, pp. 138-147, 1999.

[8] N. Pfeifer, T. Reiter, C. Briese, and W. Rieger, "Interpolation of high quality ground models from laser scanner data in forested areas," *Int. Arch. Photogramm. Remote Sens. Spatial Inf. Sci.*, vol. 32, pp. 31-36, 1999.

[9] W. Schickler and A. Thorpe, "Surface estimation based on LiDAR," *In Proceedings of the ASPRS Annual Conference*, St. Louis, Missouri, 2001.

[10] C. Briese and N. Pfeifer, "Airborne laser scanning and derivation of digital terrain models," *In Proceedings of Fifth Conference on Optical 3-D Measurement Techniques*, Vienna, Austria, 2001.

[11] G. Sithole, G. Vosselman, "Experimental comparison of filter algorithms for bare-earth extraction from airborne laser scanning point clouds," *Photogramm. Eng. Remote Sens.*, vol. 59, pp. 85-101, 2004.

[12] M. Y. Belkhouche and B. Buckles, "Iterative TIN-based automatic filtering of sparse LiDAR data," *Remote Sensing Letters*, vol. 2, no. 3, pp. 231-240, 2011.

[13] D. Mongus and B. Žalik, "Parameter-free ground filtering of LiDAR data for automatic DTM generation," *ISPRS J. Photogramm. Remote Sens.*, vol. 67, pp. 1-12, 2012.

[14] P. Lohmann, A. Koch, M. Schaeffer, "Approaches to the filtering of laser scanner data," *Int. Arch. Photogramm. Remote Sens.*, vol. 33, pp. 540-547, 2000.

[15] K. Q. Zhang, S. C. Chen, D. Whitman, M. L. Shyu, J. H. Yan, and C. C. Zhang, "A progressive morphological filter for removing nonground measurements from airborne LiDAR data," *IEEE Trans. Geosci. Remote Sens.*, vol. 41, pp. 872-882, 2003.

[16] J. Shan and A. Sampath, "Urban dem generation from raw LiDAR data: a labeling algorithm and its performance," *Photogramm. Eng. Remote Sens.*, vol. 75, pp. 427-442, 2009.

[17] X. Meng, L. Wang, J. L. Silván-Cárdenas, and N. Currit, "A multi-directional ground filtering algorithm for airborne LiDAR," *ISPRS Journal of Photogrammetry and Remote Sensing*, vol. 64, pp. 117-124, 2009.

[18] N. Haala, C. Brenner, "Extraction of Buildings and Trees in Urban Environments," *ISPRS J. Photogramm. Remote Sens.*, vol. 54, pp. 130-137, 1999.

[Email: lei.wang@uwaterloo.ca]

[19] S. Filin, “Surface clustering from airborne laser scanning data,” *Int. Arch. Photogramm. Remote Sens. Spatial Inf. Sci.*, VOL. 34, pp. 119-124, 2002.

[20] K. Jacobsen and P. Lohmann, “Segmented filtering of laser scanner dsms,” *In Proceedings of the ISPRS Working Group I/3 workshop 3-D Reconstruction From Airborne Laserscanner and InSAR Data,* Dresden, Germany, 2003.

[21] D. Tóvári and N. Pfeifer, “Segmentation based robust interpolationa new approach to laser filtering”. *Int. Arch. Photogramm. Remote Sens. Spatial Inf. Sci.*, vol. 36, pp. 79-84, 2005.

[22] T. Rabbani, F. A. van den Heuvel, and G. Vosselman, “Segmentation of point clouds using smoothness constraint,” *Int. Arch. Photogramm. Remote Sens. Spatial Inf. Sci.*, vol. 36, pp. 248-253, 2006.

[23] F. Rottensteiner, C. Baillard, G. Sohn, and M. Gerke, “ISPRS Test Project on Urban Classification And 3d Building Reconstruction,” no. 3, 2011.

[24] M. Cramer, “The DGPF test on digital aerial camera evaluation - overview and test design,” *Photogrammetrie Fernerkundung Geoinformation*, no. 2, pp. 73-82, 2010.

[Email: lei.wang@uwaterloo.ca]